\documentclass[letterpaper, 10 pt, conference]{ieeeconf}  

\IEEEoverridecommandlockouts                              
\usepackage[utf8]{inputenc}
\usepackage[T1]{fontenc}
\usepackage{graphicx}
\usepackage{amsmath}
\usepackage{amsmath,amssymb}
\usepackage{graphicx}
\usepackage{booktabs}
\usepackage{multirow}
\usepackage[caption=false,font=footnotesize]{subfig} 
\usepackage{xcolor}
\usepackage{cite}
\makeatletter
\let\NAT@parse\undefined
\makeatother

\usepackage[hidelinks]{hyperref}
\usepackage[caption=false,font=footnotesize]{subfig}

\title{\LARGE \bf
Object-Centered Reconstruction for Vision-Based 3D Force Estimation
}
\label{sec:reconstruction}

\author{
Zhonghao Zhang$^{1,*}$,
Mingyeung Wu$^{1,*}$,
Hao Yang$^{2}$,
Ayberk Acar$^{2}$,
Alan Kuntz$^{1}$,
Jie Ying Wu$^{2,\dagger}$%
\thanks{$^{*}$These authors contributed equally.}%
\thanks{$^{1}$Department of Electrical and Computer Engineering, Vanderbilt University,
Nashville, TN, United States.}%
\thanks{$^{2}$Department of Computer Science, Vanderbilt University,
Nashville, TN, United States.}%
\thanks{$^{\dagger}$Corresponding author: jieying.wu@vanderbilt.edu.}%
}

\begin{document}

\maketitle
\thispagestyle{empty}
\pagestyle{empty}

\begin{abstract}

Excessive force may damage tissue and increase the risk of anastomotic leakage in robotic colorectal surgery. Although the da Vinci 5 provides force sensing, this capability is unavailable on earlier da Vinci systems and many other surgical robotic platforms. In this work, we present a vision-based pipeline for estimating 3D interaction forces from soft-tissue deformation in stereo endoscopic video. We dynamically reconstruct the tissue point cloud in an object-centered coordinate frame, track tissue points with geometric constraints, and predict the 3D force vector with a neural network.
We progressively evaluate the pipeline on rubber-glove phantoms, \emph{ex vivo} porcine colons, and \emph{in vivo} colorectal surgical video sequences. Under varying tissue orientations and positions within the endoscopic view, as well as different camera viewpoints, the proposed method achieves average root mean square error (RMSEs) of 0.77 N and 1.30 N on the phantom and porcine colon, respectively. Compared with the camera-frame representation, the object-centered representation reduces average RMSE by 51.3\% and 56.7\%, while geometry-constrained tracking reduces RMSE by 19.8\% and 25.3\% compared with CoTracker. We further qualitatively demonstrate the feasibility of vision-based force estimation on an \emph{in vivo} colorectal surgical sequence, as a step toward clinical translation of vision-based, sensorless force estimation.

\end{abstract}

\section{Introduction}

Robot-assisted minimally invasive surgery (RMIS) provides enhanced visualization and dexterity while preserving the benefits of minimally invasive access~\cite{Moorthy2004DexterityEW}. However, surgeons using these robotic systems generally rely on visual feedback due to limited or absent force feedback, which may increase the risk of excessive tissue loading and associated complications, such as anastomotic leakage in colorectal surgery~\cite{Khalid2025,Khan2025,Turrentine2015}. Chang et al. showed that force feedback can reduce the force applied to tissue during robotic manipulation~\cite{Chang2025}. To make force sensing available to surgeons, the da Vinci 5 (Intuitive Surgical Inc., Sunnyvale, CA) offers built-in force sensing that provides surgeons with force feedback during tissue manipulation~\cite{IntuitiveDV5}. However, force sensors remain unavailable in many robotic surgical systems, and integrating dedicated sensors can be costly, motivating methods for recovering forces without direct sensing~\cite{Hosseinabadi2022}. 

In this work, we propose a stereo vision-based force estimation pipeline that reconstructs and tracks 3D tissue deformation. We represent the reconstructed tissue in an object-centered coordinate frame. We then use a spatiotemporal learning model to map the tracked deformation to a 3D interaction-force vector. During inference, this pipeline does not require force sensors and robot dynamics. Our main contributions are: 

(1) We introduce an object-centered 3D representation that describes tissue geometry independently of the global camera coordinate system. This representation significantly reduces sensitivity to changes in tissue poses, camera viewpoint, and scene configuration. 

(2) We introduce a deformation-tracking strategy for surgical scenes that are affected by specular reflections, weak texture and instrument occlusion. The method maintains a consistent 3D tissue representation by aligning visible tissue points with the current reconstruction and preserving local geometric relationships between neighboring tissue points. 

(3) We demonstrate generalization of force estimation across large changes in tissue pose, as well as qualitative results on transfer from \emph{ex vivo} porcine colon to an \emph{in vivo} colorectal surgical sequence. The model trained only on \emph{ex vivo} data produces force estimates that remain qualitatively consistent with the observed tool--tissue interaction.

\begin{figure}[!t]
    \centering

    \subfloat[]{%
        \includegraphics[
            width=0.48\columnwidth,
            trim={160 10 455 210},
            clip
        ]{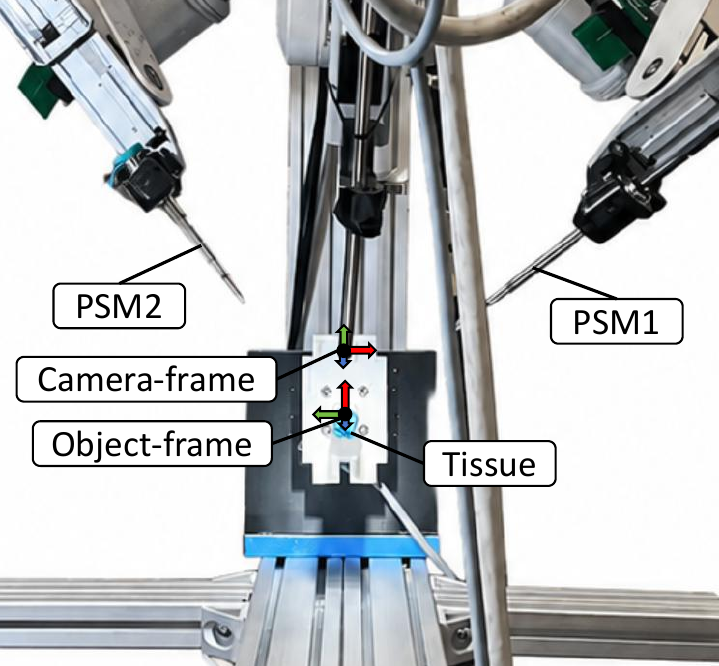}
        \label{fig:phantom_exp}
    }
    \hfill
    \subfloat[]{%
        \includegraphics[
            width=0.48\columnwidth,
            trim={40 200 550 0},
            clip
        ]{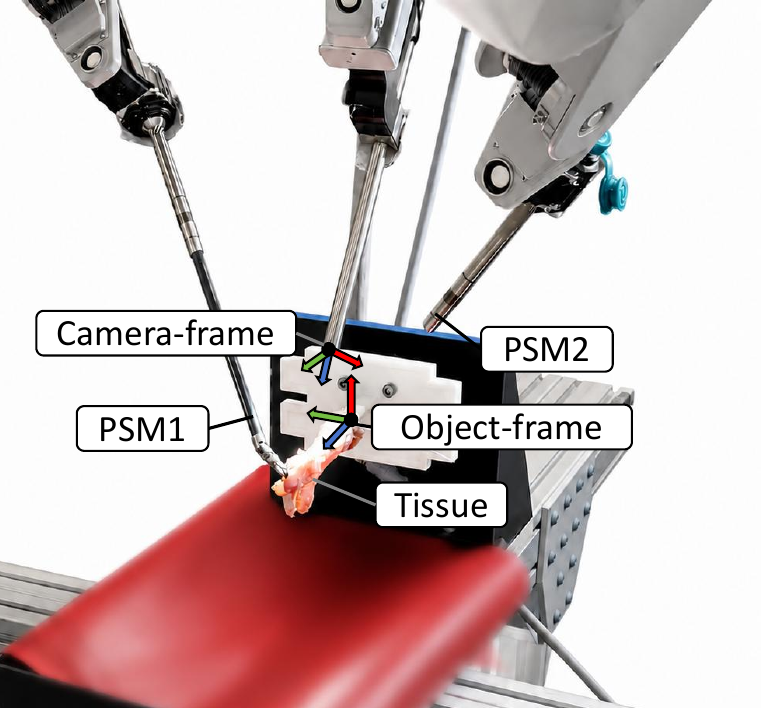}
        \label{fig:colon_exp}
    }

    \caption{Experimental setups.
    (a) Rubber-glove phantom experiment with the deformable finger segment mounted on the force/torque sensor.
    (b) \emph{Ex vivo} porcine colon experiment during robot-assisted tissue manipulation.}
    \label{fig:experiment_setup}
    \vspace{-4mm}
\end{figure}

\section{Related Works}
\begin{figure*}[t]
  \centering
  \includegraphics[
    width=\textwidth
  ]{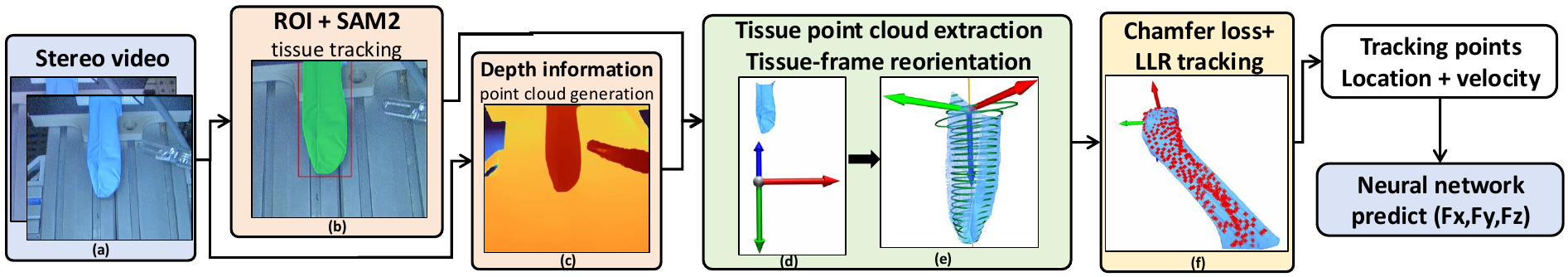}
  \caption{Visualization of the stereo reconstruction and object-centered preprocessing pipeline.
(a) Original stereo endoscopic image captured by the dVRK-Si camera.
(b) Tissue mask from SAM2, prompted by a Faster R-CNN bounding box, overlaid on the image.
(c) Depth map estimated by Stereo Any Video.
(d) Tissue point cloud reconstructed in the camera coordinate frame by combining the SAM2 segmentation mask with the estimated depth, before coordinate reorientation.
(e) Tissue point cloud reoriented from the camera frame into an object-centered coordinate frame.
(f) Tracking points overlaid on the reoriented tissue point cloud at a representative frame under stretching.}
  \label{fig:preprocessing}
  \vspace{-4mm}
\end{figure*}

Existing force estimation approaches include direct sensing and sensorless methods. Direct sensing provides force measurements through dedicated hardware, but integrating force sensors into surgical instruments can increase manufacturing costs and complicate miniaturization and sterilization~\cite{Hosseinabadi2022}. Sensorless methods instead estimate force from robot states. On the da Vinci Research Kit (dVRK) and the second-generation dVRK-Si~\cite{Kazanzides2014, DVRK_Si}, researchers have explored both model-based and learning-based approaches to solve the inverse dynamics problem of estimating external forces~\cite{Yilmaz2020,Yang2025}. However, these methods require instrument kinematics and accurate knowledge of platform-specific dynamics. Vision-based force estimation is an attractive alternative as it can infer force from tissue deformation using endoscopic video alone~\cite{Gao2018}.

Early vision-based force estimation methods measure tissue deformation from images and map it to force using explicit mechanical models. Kennedy et al. use a finite element model, while Giannarou et al. and Haouchine et al. combine stereo reconstruction with biomechanical models~\cite{Kennedy2005,Giannarou2016,Haouchine2018}. 
Other methods use neural networks to learn the relationship between tissue deformation and force. Gao et al. regress the normal contact force from RGB images and point clouds from an external RGB-D camera~\cite{Gao2018}, and Gessert et al. regress force from volumetric optical coherence tomography data~\cite{Gessert2018}. Both require imaging hardware that is rarely available in RMIS.
Some previous works combine monocular RGB images with robot kinematics or state~\cite{Chua2021,Reyzabal2024DaFoEs}. These methods are evaluated only on silicone phantoms with limited changes in camera and tissue position. Chua et al. find that vision-based networks are sensitive to viewpoint shifts, particularly along the image axis with the least depth perception~\cite{Chua2021}. Vision-only methods estimate force directly from endoscopic images without additional signals~\cite{Masui2024,Wang2026Bowel}, but monocular images lack explicit depth information. Wang et al. reconstruct 3D point clouds from stereo images, using an external structured-light projector to improve correspondence matching in challenging surgical scenes~\cite{Wang2025Structured}. This additional hardware, however, limits direct deployment with standard stereo endoscopic video. Additionally, vision-based methods generally suffer from a lack of generalization to visual scene changes and large changes in object pose~\cite{Chua2021, Yang2024}. These challenges motivate this work to use tracked features and an object-centric frame to improve generalization.


\section{Methods}
\begin{figure*}[!t]
  \centering
  \includegraphics[
    width=\textwidth
  ]{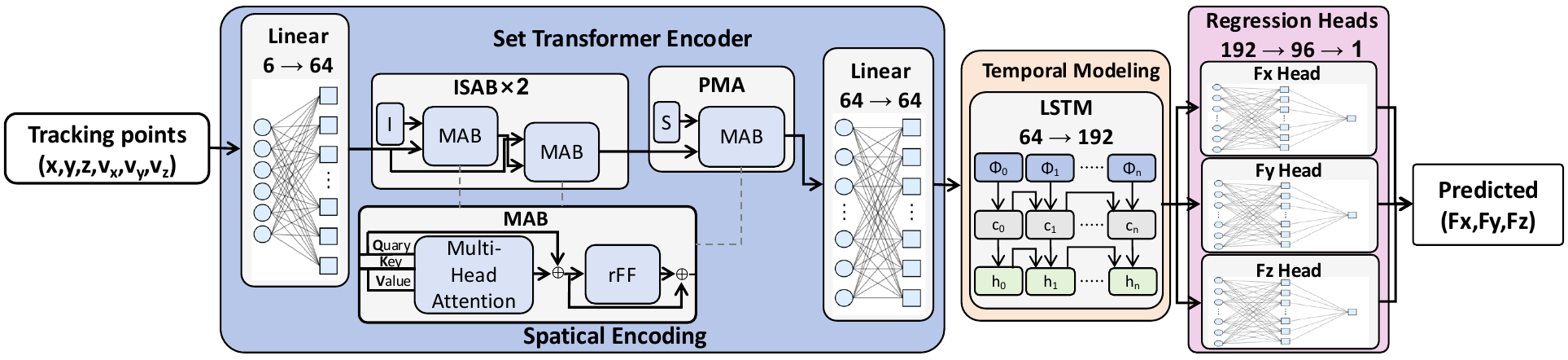}
  \caption{Architecture of the proposed model. The input is the 3D position and velocity of each tracking point. At each frame, we use a Set Transformer with two ISABs and PMA to encode the spatial relationships of tracking points. We then use an LSTM to capture the temporal information and use three independent regression heads to estimate the Cartesian force components $F_x$, $F_y$, and $F_z$.}
  \label{fig:pipeline}
  \vspace{-4mm}
\end{figure*}
Fig.~\ref{fig:preprocessing} shows the overall pipeline, which consists of three stages. First, we reconstruct the tissue in an object-centered coordinate frame, which encodes tissue geometry independently of the global camera coordinate system. Next, we track points on the reconstructed tissue to capture its deformation over time. Finally, we use a spatiotemporal neural network to estimate the 3D interaction force from the tracked points.
\subsection{Object-Centered Reconstruction}
As shown in Fig.~\ref{fig:preprocessing}(a), we record stereo endoscopic video from the dVRK-Si during tissue manipulation. Then, we use Stereo Any Video~\cite{jing2025stereoanyvideo} to estimate depth (Fig.~\ref{fig:preprocessing}(c)) and back-project the estimated depth with the calibrated camera intrinsics to reconstruct a dense 3D point cloud. In parallel, we initialize Segment Anything Model 2 (SAM2)~\cite{ravi2025sam2,Acar2026Perseus} in the first frame using a bounding-box prompt from a fine-tuned Faster R-CNN~\cite{ren2015faster} for the phantom and \emph{ex vivo} experiments (Fig.~\ref{fig:preprocessing}(b)), or manually selected prompt points for the \emph{in vivo} surgical sequence. SAM2 then propagates the tissue mask from the annotated first frame to the remainder of each sequence with no further annotation, and the resulting mask is used to extract the tissue point cloud from the dense reconstruction. 

Fig.~\ref{fig:preprocessing}(d) and (e) show the camera frame and the object frame after alignment. We assume that in the first frame of each sequence, the entire tissue is visible. We then use the complete tissue point cloud from this frame to estimate the object frame. First, we generate $36\times36$ candidate directions that cover a hemisphere. For each candidate direction $\mathbf{c}$, we cut the point cloud into slices perpendicular to candidate directions and project each slice onto its plane. We fit an ellipse to each projected slice, then map the ellipse centers back to 3D and fit a centerline with direction $\mathbf{d}$. We score each candidate by the unsigned angle between $\mathbf{c}$ and $\mathbf{d}$.
We select the $\mathbf{d}$ with the smallest unsigned angle. 

Next, we compare the ellipse sizes along the centerline to find the wider end, and we define positive $Z$ from the wider end toward the narrower end. We then project all points onto the $Z$-axis, sort the projected coordinates from $Z$ negative to $Z$ positive, and place the origin at the point corresponding to the 2nd percentile. From this origin, we align $X$ with the vertical direction(opposite to gravity) determined from the first-frame Endoscopic Camera Manipulator (ECM) pose. The ECM pose is used only to align the object-frame orientation with the ground-truth force coordinate frame and is not provided to the neural network as an input during training. $Y$ follows from the right-hand rule. Fig.~\ref{fig:experiment_setup} illustrates the relationship between the camera frame and the object-centered frame. In the resulting object frame, the origin lies approximately at the tissue attachment point rather than exactly at the force-sensor coordinate origin.

\subsection{Geometry-Constrained 3D Point Tracking}
We use the first frame of each sequence as the reference frame, where the tissue region is fully visible. From this frame, we reconstruct the tissue point cloud and initialize a fixed set of 256 3D tracking points.
At frame $t$, let $\mathcal{P}_t$ denote the tracking points and $\mathcal{Q}_t$ the reconstructed visible tissue point cloud. 
For each subsequent frame, we track the tissue points in 3D and use the 2D SAM2 mask to determine point visibility. We combine the SAM2 segmentation mask with the calibrated camera parameters to determine whether each 3D tracking point projects inside the current tissue mask. We classify a tracking point as visible only if it falls inside the current SAM2 tissue mask, and denote the resulting subset as $\mathcal{P}_t^{\mathrm{vis}} \subseteq \mathcal{P}_t$. We model the displacement of each tracking point with a sinusoidal representation network (SIREN)~\cite{sitzmann2020implicit} and optimize its parameters with Adam~\cite{kingma2015adam}, initialized from the solution of the previous frame. We use the Chamfer loss to measure the distance between the visible tracking points and the reconstructed tissue point cloud,
\begin{equation}
L_{\mathrm{Chamfer}}
=
\frac{1}{|\mathcal{P}_t^{\mathrm{vis}}|}
\sum_{\mathbf{p}_i^t \in \mathcal{P}_t^{\mathrm{vis}}}
\min_{\mathbf{q}_j^t \in \mathcal{Q}_t}
\left\|
\mathbf{p}_i^t-\mathbf{q}_j^t
\right\|_2^2,
\end{equation}
where $\mathbf{p}_i^t$ denotes the $i$-th visible tracking point and $\mathbf{q}_j^t$ denotes a point in the reconstructed visible tissue point cloud.

We use the local linear reconstruction (LLR) loss to measure how well each tracking point is reconstructed from its neighbors,
\begin{equation}
L_{\mathrm{LLR}}^{k}
=
\sum_{\mathbf{p}_i^t \in \mathcal{P}_t}
\left\|
\mathbf{p}_i^t - \sum_{j\in\mathcal{N}^{k}(i)} w_{ij}^{k}\mathbf{p}_j^t
\right\|_2^2,
\end{equation}
where $\mathcal{N}^{k}(i)$ and $w_{ij}^{k}$ are the neighbors and reconstruction weights of point $i$. We compute $\mathcal{N}^{\mathrm{ref}}$ and $w^{\mathrm{ref}}$ once from the reference reconstruction to preserve the tissue's original local structure. We recompute $\mathcal{N}^{\mathrm{prev}}$ and $w^{\mathrm{prev}}$ at every frame to keep the tracking temporally consistent. We jointly update the complete set of tracking points by minimizing
\begin{equation}
L =
\lambda_{\mathrm{Chamfer}}L_{\mathrm{Chamfer}}
+
\lambda_{\mathrm{ref}}L_{\mathrm{LLR}}^{\mathrm{ref}}
+
\lambda_{\mathrm{prev}}L_{\mathrm{LLR}}^{\mathrm{prev}}
\end{equation}
Fig.~\ref{fig:preprocessing}(f) shows the tracked points lifted onto the reconstructed 3D point cloud.
\subsection{Spatiotemporal Force Regression}
As shown in Fig.~\ref{fig:pipeline}, after obtaining the tracked 3D point positions at each frame, we first convert the spatial coordinates to millimeters and approximate its velocities using a central difference scaled by \(\mathbf{v}_t^{i} = \frac{f_s}{2}(\mathbf{p}_{t+1}^{i} - \mathbf{p}_{t-1}^{i})\), where \(f_s\) is the sampling rate, \(\mathbf{p}_t^{i}\in\mathbb{R}^{3}\) denotes the 3D position of the \(i\)-th tracked point at frame \(t\), and \(\mathbf{v}_t^{i}\in\mathbb{R}^{3}\) denotes its estimated velocity. We then concatenate the position and velocity to form a six-dimensional point token.

Next, we encode the point tokens of each frame with a Set Transformer~\cite{lee2019set}, as shown in Fig.~\ref{fig:pipeline}. We project each token to 64 dimensions, pass it through two induced set attention blocks (ISABs) and aggregate the point features into a frame-level embedding $\mathbf{z}_t \in \mathbb{R}^{64}$ with a pooling by multihead attention (PMA) block. Both ISAB and PMA are constructed using multihead attention blocks (MABs), as illustrated in Fig.~\ref{fig:pipeline}.
To capture the temporal relationship between frames, a single-layer unidirectional long short-term memory (LSTM) network~\cite{hochreiter1997long} with 192 hidden units processes overlapping clips of \(T=16\) frames with a stride of 4. Three independent regression heads then map the LSTM feature at each frame to \(F_x\), \(F_y\), and \(F_z\).
\section{Experimental Setup and Evaluation}
\setlength{\textfloatsep}{1pt}
\begin{figure}[!b]
    \centering

    \subfloat[]{%
        \includegraphics[
            width=0.48\columnwidth
        ]{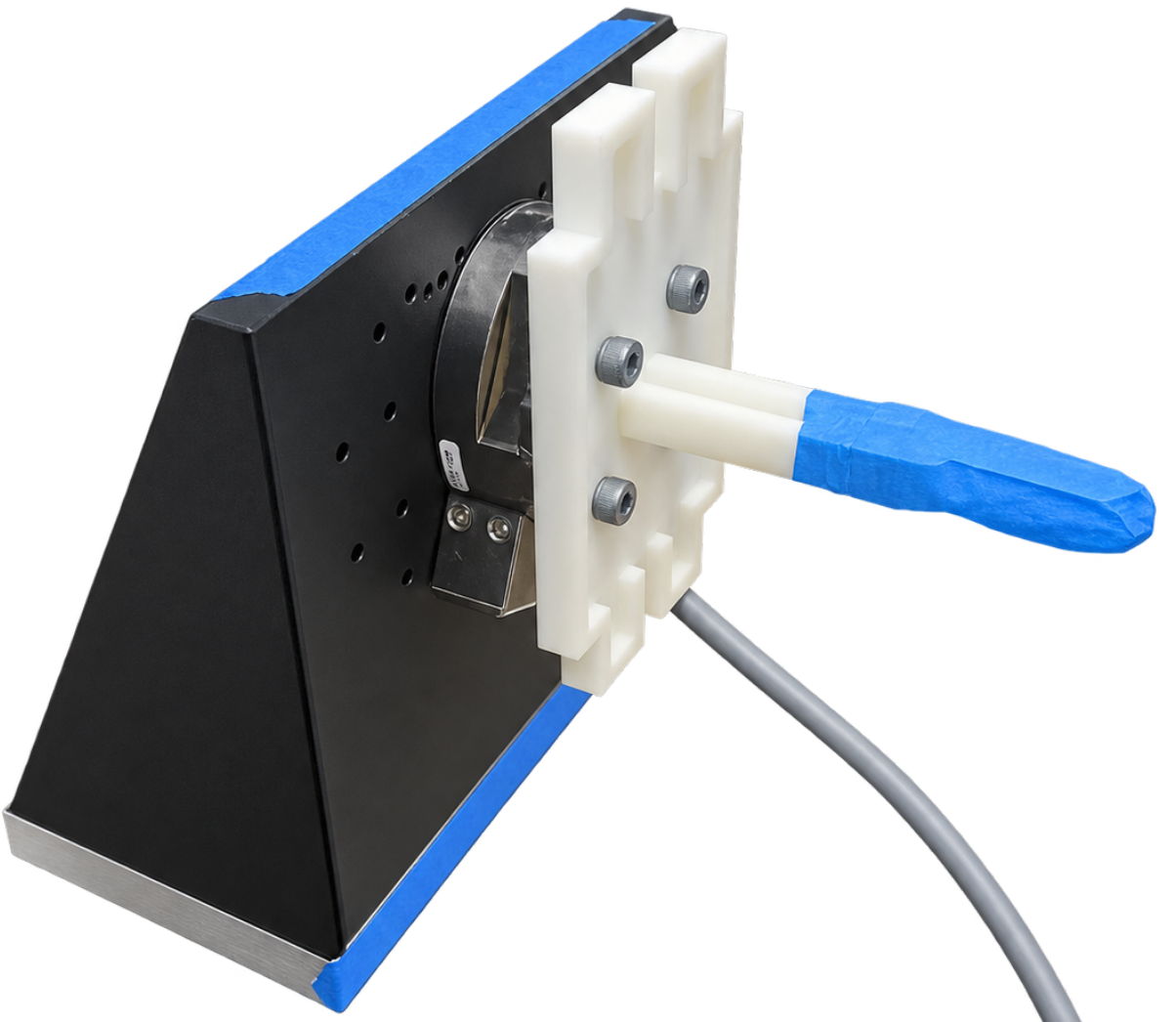}
        \label{fig:tissue_sensor}
    }
    \hfill
    \subfloat[]{%
        \includegraphics[
            width=0.41\columnwidth
        ]{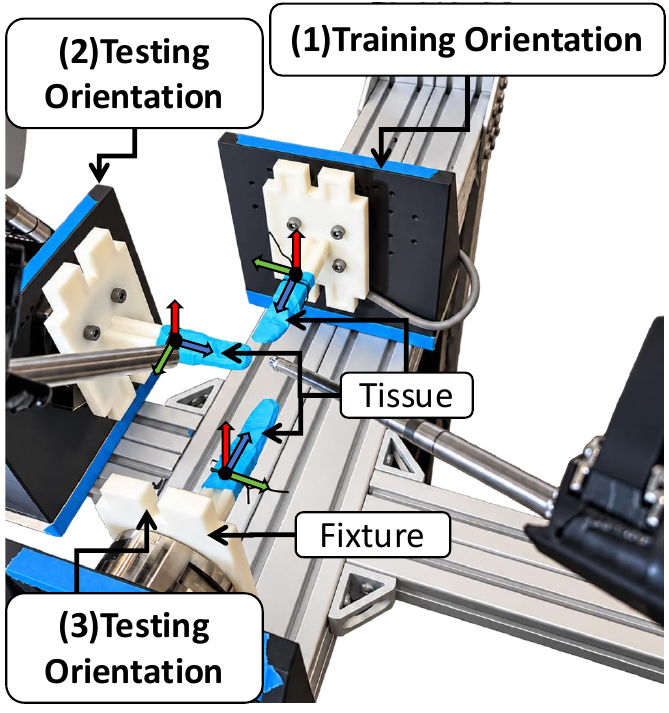}
    }

    \caption{
    Tissue fixture and experimental orientations.
(a) Tissue mounted on the force/torque sensor; the tissue–sensor assembly rotates together.
(b) Overlay of the three tissue orientations used in the experiments. Training/validation orientation (1) and test orientations (2) and (3), rotated by approximately \(90^\circ\) and \(180^\circ\) in the horizontal plane.}
    \label{fig:fixture_setup}
\end{figure}
For both studies, we used dVRK-Si teleoperation and recorded synchronized stereo endoscopic video and force measurements at 30 Hz. An ATI Gamma six-axis force/torque sensor (ATI Industrial Automation, Apex, NC, USA) provided the ground-truth interaction forces. As shown in Fig.~\ref{fig:fixture_setup}(a), the tissue and force sensor are mounted on the same fixture and rotate together; therefore, the ground-truth force coordinate frame rotates with the tissue. The sensor measures forces transmitted from the tissue through the fixture.

We train a separate model for each study (phantom and \emph{ex vivo} colon) while keeping the training configuration identical. We train the randomly initialized network with the mean squared error between the predicted and measured forces. We use AdamW with a learning rate of \(5\times10^{-4}\) and a weight decay of \(10^{-3}\), and train for 60 epochs. We select the checkpoint with the lowest validation loss. The training and validation sets are collected under the same object orientation, while the held-out test sets contain different tissue and ECM poses. We do not use the test sets for model selection. For the test experiments, the glove phantom and colon are rotated counterclockwise (viewed from above) by approximately \(90^\circ\) and \(180^\circ\) in the horizontal plane (Fig.~\ref{fig:fixture_setup}(b)). We also change the ECM viewpoint and the object location in the image.
\subsection{Rubber-Glove Phantom Study}
As shown in Fig.~\ref{fig:experiment_setup}(a), we rigidly mounted the rubber-glove phantom to the force/torque sensor and stretched it along multiple directions using the dVRK-Si. We used three gloves from the same box during data collection.

The dataset contains 16,301 frames (approximately 9 min of recordings) in total. 11,267 frames were used for training, 3735 for validation, and 1299 for testing, with a training-validation-test split ratio of around 7:2:1. The 1299 frames for the test sequences were collected under different geometric and viewpoint conditions, with substantial repositioning within the endoscopic view. These changes alter the tissue appearance, depth, and occlusion in the camera view, while the force distribution remains comparable to that of the training data (Fig.~\ref{fig:force_distribution}(a)).
\subsection{Ex Vivo Porcine Colon Study}

\begin{figure}[t]
    \centering

    \subfloat[]{%
        \includegraphics[
            width=\columnwidth,
            trim={0 0 0 0},
            clip
        ]{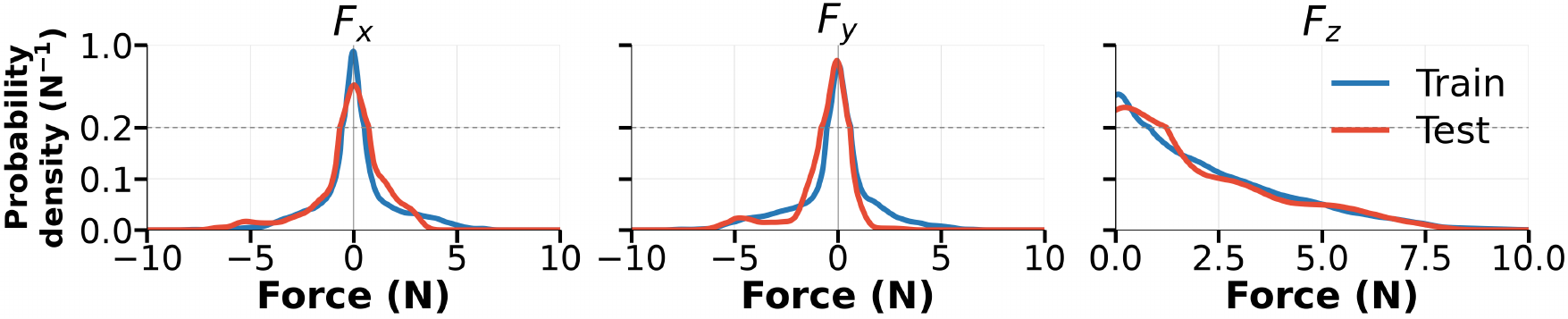}
        \label{fig:force_distribution_a}
    }

    \vspace{0.05cm}

    \subfloat[]{%
        \includegraphics[
            width=\columnwidth,
            trim={0 0 0 0},
            clip
        ]{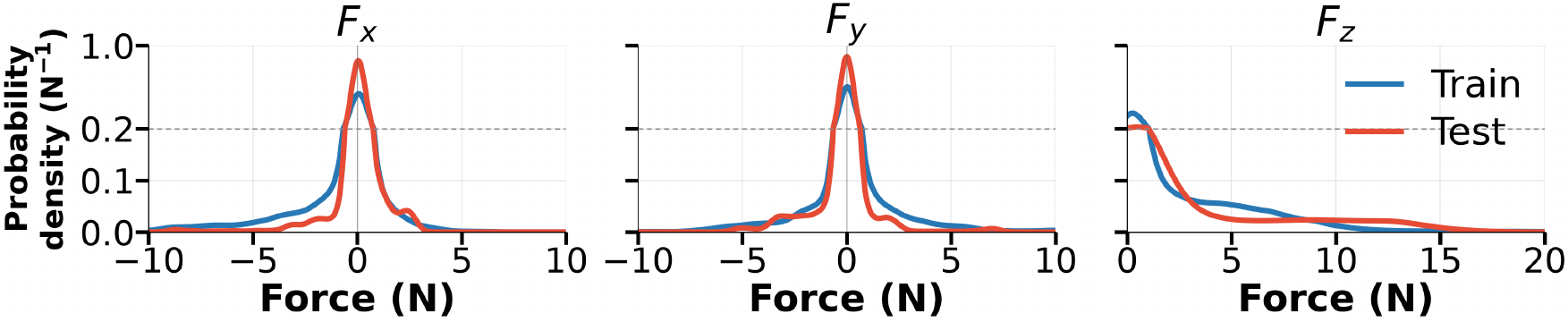}
        \label{fig:force_distribution_b}
    }

\caption{
Comparison of the measured Cartesian force distributions between the training and test sets. (a) Rubber-glove phantom study. (b) \textit{Ex vivo} porcine colon study. Each panel shows kernel density estimates of $F_x$, $F_y$, and $F_z$. We compress the vertical axis between 0.2 and 1 for visualization.}
    \label{fig:force_distribution}
\end{figure}
For the \emph{ex vivo} experiment, we attached one end of the porcine colon to the force/torque sensor, as shown in Fig.~\ref{fig:experiment_setup}(b), and pulled the tissue along multiple directions using the dVRK-Si. We used two samples cut from one colon section during data collection.

This dataset contains 5,283 frames (approximately 3 min), including 3540 frames for training, 898 for validation, and 845 testing, with a split ratio of around 7 : 1.5 : 1.5.  We varied the tissue configuration by repositioning and rotating the colon by $90^\circ$ and $180^\circ$, placing it at different locations within the endoscopic view, and changing the ECM viewpoint. The resulting force distribution is shown in Fig.~\ref{fig:force_distribution}(b).
\subsection{Qualitative In Vivo Evaluation}

To examine transfer beyond the controlled phantom and colon experiments, we further apply the model trained on \emph{ex vivo} data to \emph{in vivo} recordings from human colorectal surgery. This experiment introduces a substantially different visual and mechanical domain. Since no ECM pose or force sensor is available, the $X$-axis of the object frame is chosen arbitrarily in the plane perpendicular to $Z$, serving only as a reference for expressing the estimated force direction. No \emph{in vivo} sequence is used during model training or model selection.
\subsection{Ablation and Baseline Studies}

\textbf{Ablation studies:} For both the glove phantom and \emph{ex vivo} colon experiments, we evaluate the proposed method using the following ablation settings. Each setting below modifies only the corresponding component of the proposed pipeline.

\emph{(1) Camera frame:}
We remove the object-frame alignment and directly represent the 256 tracking points in the camera frame.

\emph{(2) CoTracker:}
We replace the proposed geometry-constrained point tracking with CoTracker.

\emph{(3) Point cloud:}
We remove the tracking-point representation and directly use the reconstructed tissue point cloud after object-frame alignment and tissue-region segmentation as input. Since point correspondences are not maintained across frames, no point-wise velocity is used.

\emph{(4) Position only:}
We remove the point velocity and use only the 3D tracking-point positions as input.

\emph{(5) Velocity only:}
We remove the point positions and use only the 3D tracking-point velocities as input.

\emph{(6) Top/Bottom 20\%:}
We sort the tracking points by their $z$ coordinates and use the highest 20\% and lowest 20\% of the points for force estimation.

\textbf{Baseline comparisons:} For baseline comparison, we further evaluate the following image-based methods on both experimental settings:

\emph{(1) RGB only:}
We use only monocular RGB image sequences as input and apply a pretrained 3D ResNet-18 to directly estimate the 3D force~\cite{Wang2026Bowel}.

\emph{(2) Masked  Depth + ViT-Small:}
We encode the SAM2-masked depth image with a pretrained ViT-Small and feed the resulting features into the same LSTM and regression heads to estimate the 3D force.
\begin{figure}[t]
    \centering
    \subfloat[]{%
        \includegraphics[
            width=0.30\columnwidth,
            trim=105 0 20 0,
            clip
        ]{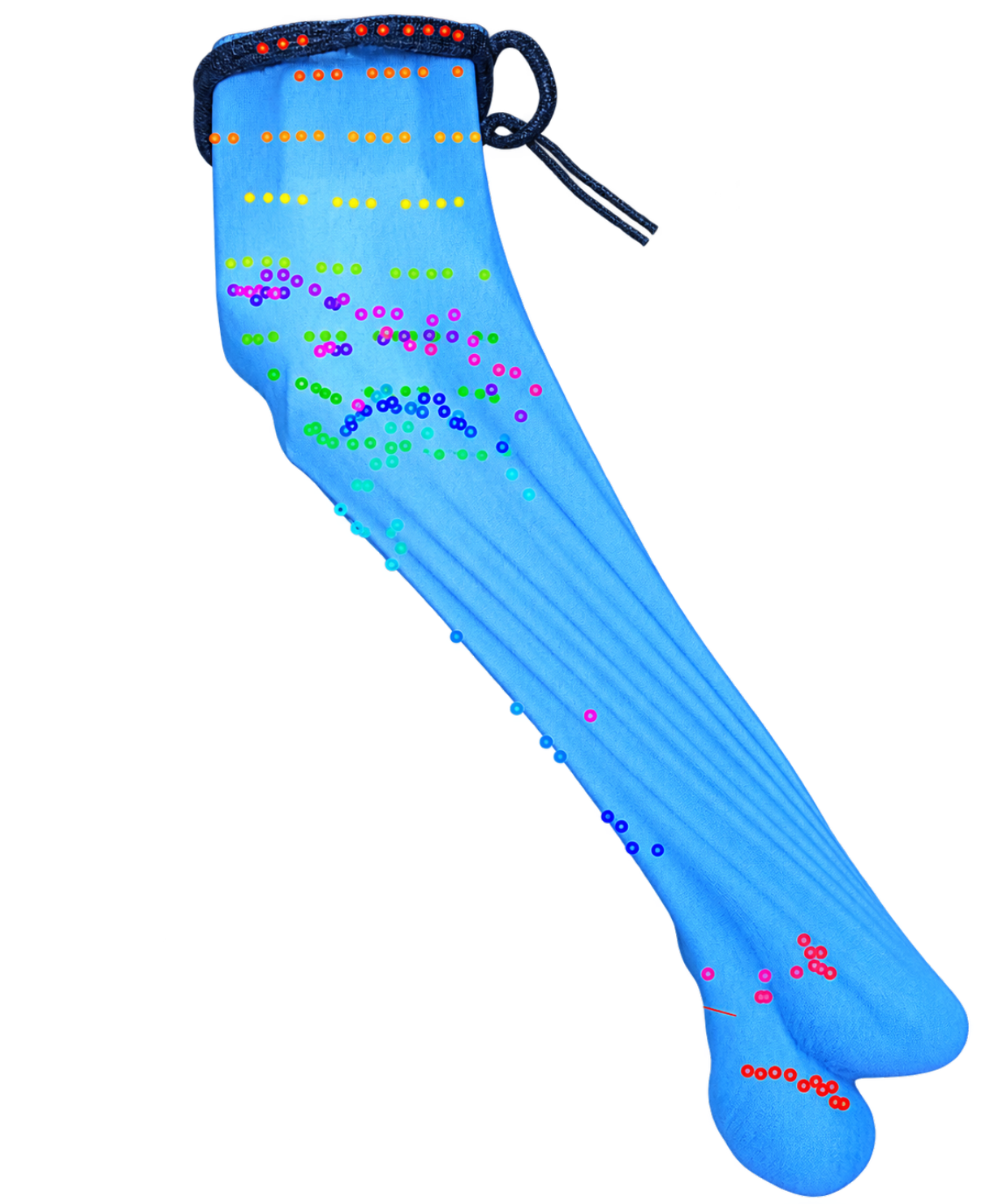}
    }
    \hfill
    \subfloat[]{%
        \includegraphics[
            width=0.30\columnwidth,
            trim=170 0 510 0,
            clip
        ]{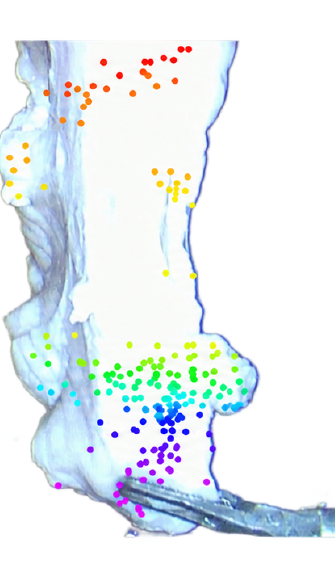}
    }
    \hfill
    \subfloat[]{%
        \includegraphics[
            width=0.30\columnwidth,
            trim=580 0 100 0,
            clip
        ]{figures/cotracker_Oerag.pdf}
    }
    \caption{
    Comparison of point tracking performance.
    (a) CoTracker tracking results on the rubber-glove phantom.
    (b) CoTracker tracking results on the ex vivo colon after temporal propagation.
    (c) Tracking results of our proposed method on the ex vivo colon after temporal propagation.
    }
    \label{fig:cotracker}
\end{figure}
\begin{figure*}[!t]
    \centering

    \subfloat[]{%
        \includegraphics[
            width=0.49\textwidth,
            trim={0 15 0 25},
            clip
        ]{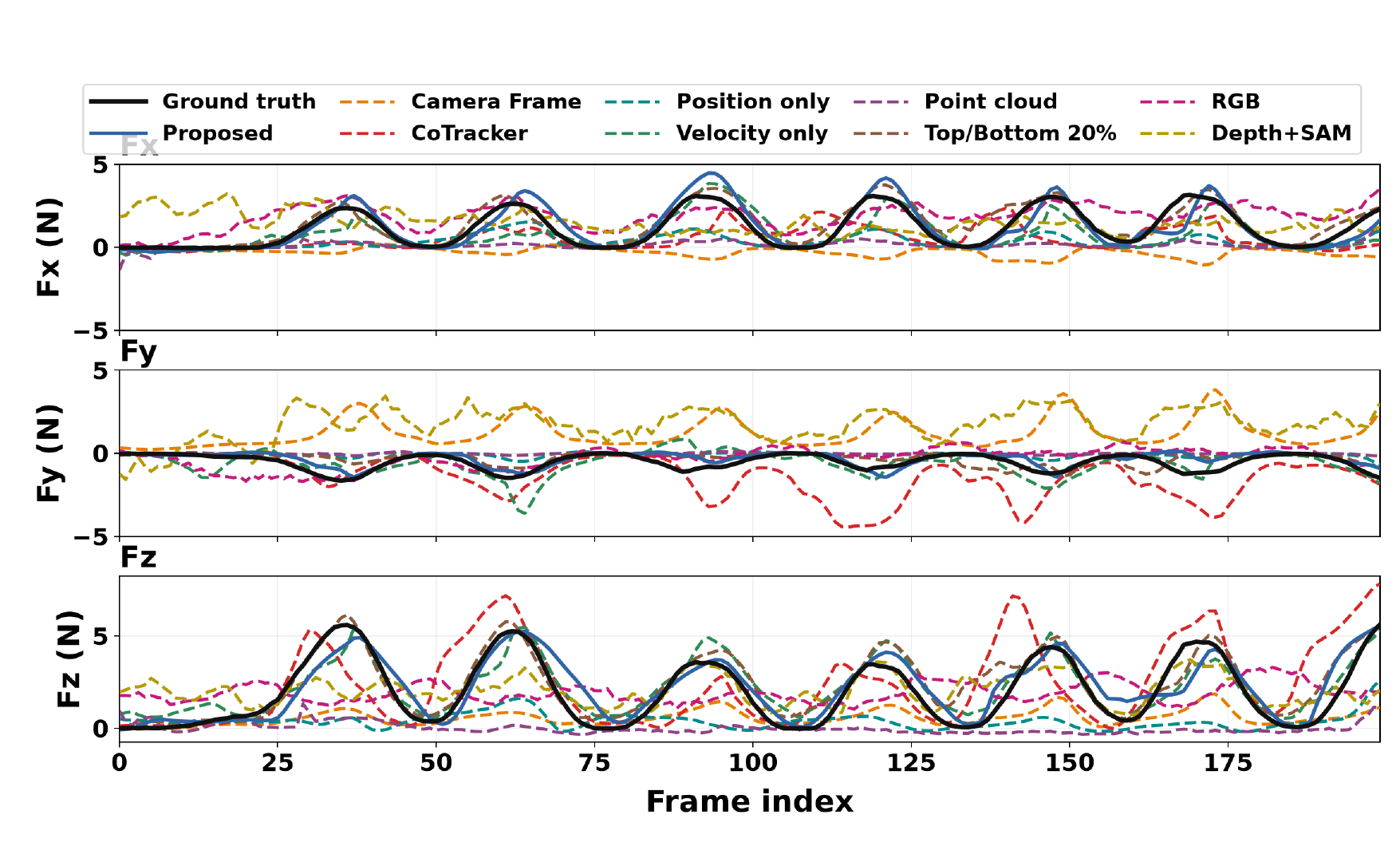}
        \label{fig:phantom_curve}
    }
    \hfill
    \subfloat[]{%
        \includegraphics[
            width=0.49\textwidth,
            trim={0 15 0 25},
            clip
        ]{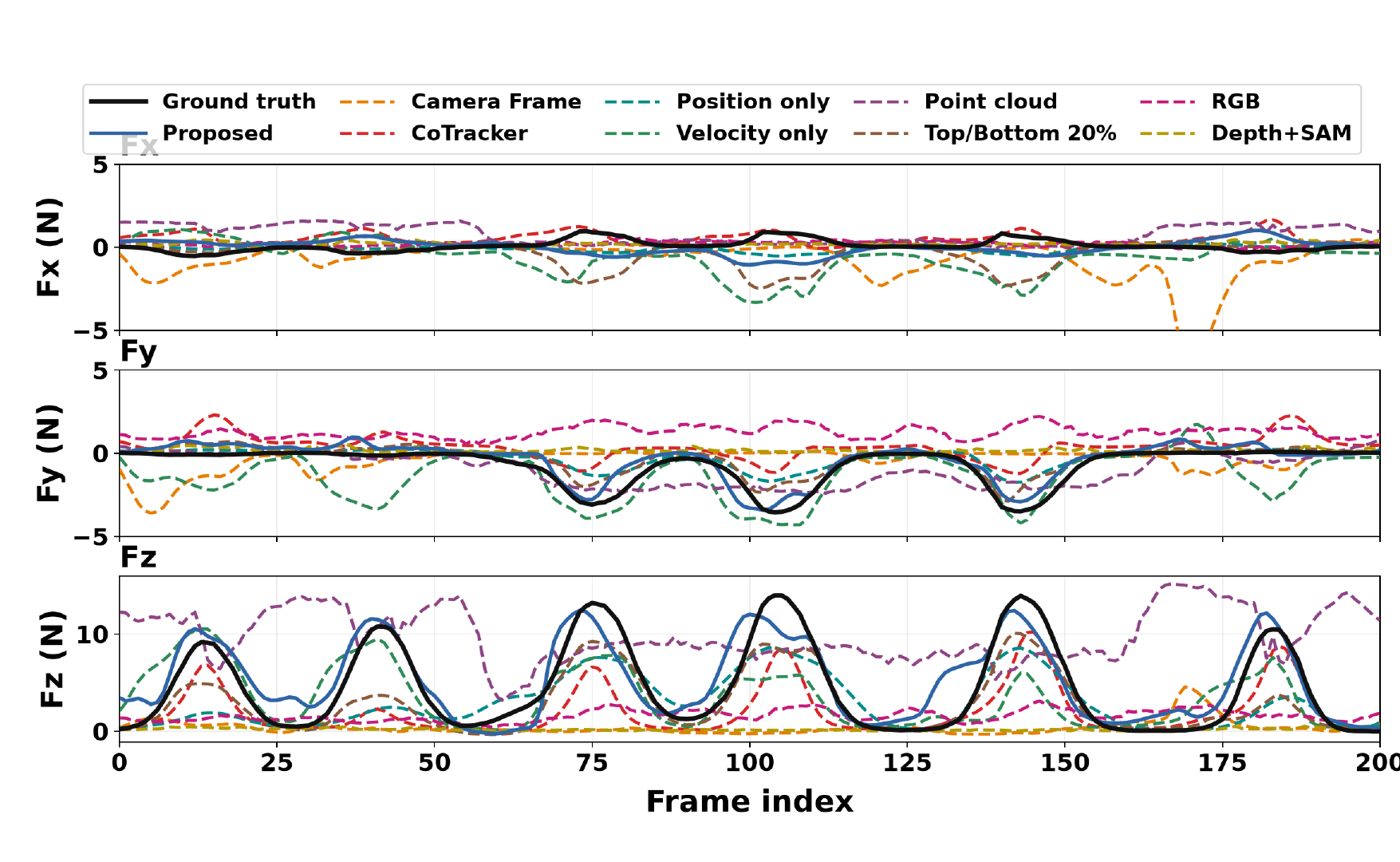}
        \label{fig:colon_curve}
    }

    \caption{
    Comparison between the predicted and ground-truth interaction forces measured by the force sensor.
(a) Rubber-glove phantom study.
(b) ex vivo porcine colon study.
    }
    \label{fig:force_curve}
\end{figure*}

\begin{figure*}[!t]
    \centering
    \includegraphics[
        width=0.98\textwidth,
        trim=20 200 20 20,
        clip
    ]{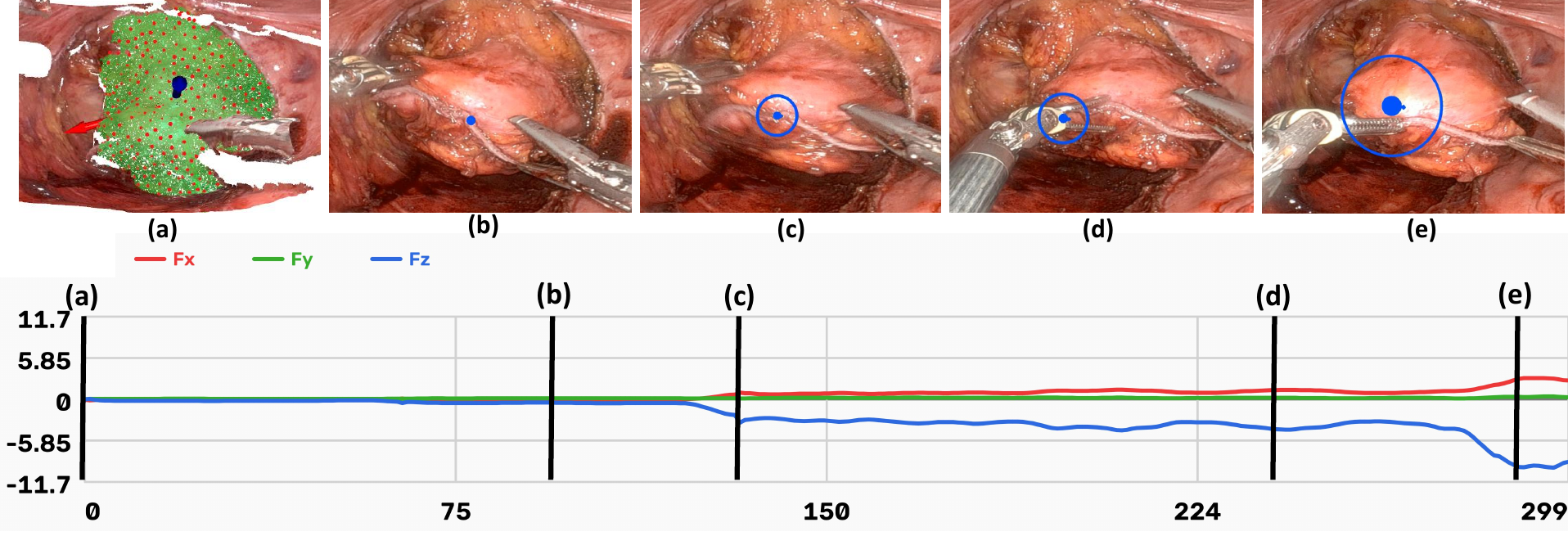}
    \caption{\emph{In vivo} force estimation during a single tissue-pulling motion. The estimated force is transformed back into the camera frame for visualization. (a) Zero-force frame: reconstructed point cloud overlaid with the SAM2 mask and tracking points. (b)--(e) Endoscopic frames at four force levels; the blue circle indicates the force component along the camera depth direction, and the arrow indicates the in-plane component. The plot below shows the estimated force (N) versus frame number over the entire video sequence, sampled at 50 Hz, and the marked points correspond to the frames shown in (a)--(e), respectively. The force magnitude shown on the y-axis is provided only as a relative reference and does not represent an absolute calibrated force measurement.}
    \label{fig:invivo_validation}
\end{figure*}

\section{Results}
We quantify force estimation accuracy using root mean square error (RMSE), normalized root mean square error (NRMSE), and the coefficient of determination ($R^2$). We compute NRMSE by dividing each force component's RMSE by its corresponding ground-truth force range in the test set.

To evaluate tracking quality during tissue interaction, we first identified local force magnitude peaks using a minimum prominence of 2.0 N and a separation of 15 frames. At each peak frame, we divided the SAM2 tissue mask into 100 regions of approximately equal pixel area and defined the coverage rate as the fraction of regions containing at least one tracking point. We also measured clustering with a normalized nearest-neighbor index (NNI), which divides the mean nearest-neighbor distance between tracking points at the peak frame by its value at the first frame, where the tissue is undeformed. We expect a slightly higher value at peak frames because the tissue is stretched. Table~\ref{tab:tracking_quality} reports the average coverage and NNI at peak frames.

\begin{table}[t]
\centering
\caption{Tracking quality at peak-force frames for the rubber-glove phantom and \textit{ex vivo} colon. Coverage measures the spatial distribution of tracking points over the tissue, and NNI measures point clustering, with lower values indicating more severe clustering.}
\label{tab:tracking_quality}
\small
\setlength{\tabcolsep}{3.5pt}
\begin{tabular}{@{}lcccc@{}}
\toprule
& \multicolumn{2}{c}{\textbf{Rubber-glove phantom}}
& \multicolumn{2}{c}{\textbf{\textit{Ex vivo} colon}} \\
\cmidrule(lr){2-3}\cmidrule(lr){4-5}
\textbf{Method}
& \textbf{Coverage (\%)}
& \textbf{NNI}
& \textbf{Coverage (\%)}
& \textbf{NNI} \\
\midrule
CoTracker
& 69.6 & 0.90
& 68.8 & 0.93 \\

Proposed
& \textbf{76.5} & \textbf{1.08}
& \textbf{69.4} & \textbf{1.01} \\
\bottomrule
\end{tabular}
\end{table}

\begin{table*}[t]
\centering
\caption{Force estimation performance for the rubber-glove phantom and ex vivo porcine colon experiments. RMSE is reported in Newtons (N). We compute means over all test frames pooled from both test orientations; $\pm$ denotes the standard deviation between orientations.}
\label{tab:force_results}

\scriptsize
\renewcommand{\arraystretch}{0.95}
\setlength{\tabcolsep}{1.8pt}

\begin{tabular*}{\textwidth}
{@{\extracolsep{\fill}}lcccccccc@{}}
\hline

\textbf{Method} &
\multicolumn{4}{c}{\textbf{RMSE (N)}} &
\multicolumn{4}{c}{\textbf{NRMSE (\%)}} \\

\cline{2-5}
\cline{6-9}

&
$\mathbf{F_x}$ &
$\mathbf{F_y}$ &
$\mathbf{F_z}$ &
\textbf{Avg.} &
$\mathbf{F_x}$ &
$\mathbf{F_y}$ &
$\mathbf{F_z}$ &
\textbf{Avg.} \\

\hline

\multicolumn{9}{l}{\textbf{Rubber-glove phantom}} \\

Proposed
& $0.91 \pm 0.24$
& $\mathbf{0.61 \pm 0.54}$
& $\mathbf{0.78 \pm 0.05}$
& $\mathbf{0.77 \pm 0.28}$
& $9.29 \pm 2.42$
& $\mathbf{6.24 \pm 5.55}$
& $\mathbf{10.17 \pm 0.66}$
& $\mathbf{8.57 \pm 2.88}$ \\

Camera Frame
& $1.31 \pm 0.14$
& $1.27 \pm 0.60$
& $2.17 \pm 0.39$
& $1.58 \pm 0.37$
& $13.27 \pm 1.41$
& $13.11 \pm 6.14$
& $28.40 \pm 5.07$
& $18.26 \pm 0.11$ \\

CoTracker
& $\mathbf{0.88 \pm 0.17}$
& $0.73 \pm 0.45$
& $1.28 \pm 0.34$
& $0.96 \pm 0.32$
& $\mathbf{8.97 \pm 1.68}$
& $7.54 \pm 4.63$
& $16.72 \pm 4.44$
& $11.07 \pm 0.62$ \\

Position only
& $1.20 \pm 0.68$
& $1.04 \pm 0.73$
& $1.94 \pm 0.58$
& $1.39 \pm 0.67$
& $12.21 \pm 6.93$
& $10.70 \pm 7.54$
& $25.36 \pm 7.59$
& $16.09 \pm 7.35$ \\

Velocity only
& $1.39 \pm 0.18$
& $1.35 \pm 1.29$
& $1.24 \pm 0.47$
& $1.32 \pm 0.64$
& $14.12 \pm 1.79$
& $13.87 \pm 13.29$
& $16.15 \pm 6.07$
& $14.72 \pm 7.05$ \\

Point cloud
& $0.98 \pm 0.35$
& $0.78 \pm 0.50$
& $1.61 \pm 0.11$
& $1.12 \pm 0.32$
& $9.92 \pm 3.53$
& $8.07 \pm 5.13$
& $21.08 \pm 1.49$
& $13.02 \pm 3.38$ \\

Top/Bottom 20\%
& $0.91 \pm 0.10$
& $1.17 \pm 0.85$
& $1.29 \pm 0.55$
& $1.12 \pm 0.50$
& $9.21 \pm 1.04$
& $12.01 \pm 8.71$
& $16.89 \pm 7.21$
& $12.70 \pm 5.65$ \\

RGB only
& $1.37 \pm 0.42$
& $1.55 \pm 0.70$
& $1.89 \pm 0.25$
& $1.60 \pm 0.46$
& $13.88 \pm 4.27$
& $16.00 \pm 7.23$
& $24.68 \pm 3.30$
& $18.19 \pm 0.12$ \\

Masked Depth + ViT-Small
& $2.40 \pm 0.68$
& $1.60 \pm 0.50$
& $1.90 \pm 0.28$
& $1.97 \pm 0.49$
& $24.39 \pm 6.92$
& $16.48 \pm 5.13$
& $24.87 \pm 3.69$
& $21.91 \pm 2.79$ \\

\hline

\multicolumn{9}{l}{\textbf{Ex vivo porcine colon}} \\

Proposed
& $1.04 \pm 0.11$
& $\mathbf{0.63 \pm 0.19}$
& $\mathbf{2.23 \pm 0.13}$
& $\mathbf{1.30 \pm 0.14}$
& $8.94 \pm 0.94$
& $\mathbf{4.98 \pm 1.47}$
& $\mathbf{13.44 \pm 0.80}$
& $\mathbf{9.12 \pm 0.44}$ \\

Camera Frame
& $1.50 \pm 0.63$
& $2.48 \pm 0.32$
& $5.02 \pm 2.12$
& $3.00 \pm 1.02$
& $12.88 \pm 5.38$
& $19.70 \pm 2.53$
& $30.33 \pm 12.79$
& $20.97 \pm 5.21$ \\

CoTracker
& $1.08 \pm 0.09$
& $1.37 \pm 0.29$
& $2.77 \pm 0.30$
& $1.74 \pm 0.23$
& $9.28 \pm 0.79$
& $10.89 \pm 2.32$
& $16.74 \pm 1.80$
& $12.30 \pm 0.44$ \\

Position only
& $\mathbf{1.02 \pm 0.20}$
& $1.00 \pm 0.44$
& $3.19 \pm 0.17$
& $1.74 \pm 0.27$
& $\mathbf{8.73 \pm 1.74}$
& $7.96 \pm 3.52$
& $19.29 \pm 1.00$
& $11.99 \pm 0.26$ \\

Velocity only
& $1.65 \pm 0.64$
& $1.63 \pm 0.17$
& $2.77 \pm 0.12$
& $2.02 \pm 0.31$
& $14.14 \pm 5.48$
& $12.97 \pm 1.34$
& $16.74 \pm 0.71$
& $14.62 \pm 1.62$ \\

Point cloud
& $1.61 \pm 0.56$
& $1.87 \pm 0.08$
& $5.71 \pm 2.43$
& $3.07 \pm 1.02$
& $13.82 \pm 4.78$
& $14.85 \pm 0.67$
& $34.48 \pm 14.66$
& $21.05 \pm 6.26$ \\

Top/Bottom 20\%
& $1.21 \pm 0.27$
& $0.72 \pm 0.22$
& $2.83 \pm 0.22$
& $1.59 \pm 0.24$
& $10.38 \pm 2.28$
& $5.71 \pm 1.71$
& $17.12 \pm 1.35$
& $11.07 \pm 0.64$ \\

RGB only
& $1.24 \pm 0.26$
& $2.11 \pm 0.58$
& $4.86 \pm 1.49$
& $2.74 \pm 0.78$
& $10.64 \pm 2.24$
& $16.72 \pm 4.64$
& $29.35 \pm 9.01$
& $18.90 \pm 2.21$ \\

Masked Depth + ViT-Small
& $1.59 \pm 0.49$
& $1.42 \pm 0.51$
& $5.03 \pm 1.50$
& $2.68 \pm 0.83$
& $13.64 \pm 4.16$
& $11.23 \pm 4.03$
& $30.40 \pm 9.06$
& $18.42 \pm 3.06$ \\

\hline
\end{tabular*}
\end{table*}

\begin{table*}[t]
\centering
\caption{$R^2$ performance for the rubber-glove phantom and
ex vivo porcine colon experiments. We compute means over all test frames pooled from both test orientations; $\pm$ denotes the standard deviation between orientations.}
\label{tab:r2_results}

\scriptsize
\renewcommand{\arraystretch}{1.05}
\setlength{\tabcolsep}{3.5pt}

\begin{tabular*}{\textwidth}{
@{\extracolsep{\fill}}
l
cccc
cccc
@{}
}
\hline

\multirow{2}{*}{\textbf{Method}} &
\multicolumn{4}{c}{\textbf{Rubber-glove phantom}} &
\multicolumn{4}{c}{\textbf{Ex vivo porcine colon}} \\

\cline{2-5}
\cline{6-9}

&
$\mathbf{F_x}$ &
$\mathbf{F_y}$ &
$\mathbf{F_z}$ &
\textbf{Avg.} &
$\mathbf{F_x}$ &
$\mathbf{F_y}$ &
$\mathbf{F_z}$ &
\textbf{Avg.} \\

\hline

Proposed
& $0.60 \pm 0.86$
& $\mathbf{0.72 \pm 0.06}$
& $\mathbf{0.82 \pm 0.10}$
& $\mathbf{0.71 \pm 0.34}$
& $0.25 \pm 0.58$
& $\mathbf{0.74 \pm 0.11}$
& $\mathbf{0.71 \pm 0.18}$
& $\mathbf{0.57 \pm 0.29}$ \\

Camera Frame
& $0.18 \pm 0.57$
& $-0.23 \pm 2.65$
& $-0.38 \pm 0.05$
& $-0.14 \pm 1.09$
& $-0.56 \pm 0.37$
& $-3.10 \pm 3.00$
& $-0.47 \pm 0.76$
& $-1.37 \pm 1.38$ \\

CoTracker
& $\mathbf{0.63 \pm 0.69}$
& $0.59 \pm 0.59$
& $0.52 \pm 0.11$
& $0.58 \pm 0.46$
& $0.19 \pm 0.68$
& $-0.25 \pm 1.62$
& $0.55 \pm 0.34$
& $0.16 \pm 0.88$ \\

Position only
& $0.31 \pm 2.42$
& $0.18 \pm 0.83$
& $-0.10 \pm 1.35$
& $0.13 \pm 1.54$
& $\mathbf{0.28 \pm 0.30}$
& $0.33 \pm 0.09$
& $0.41 \pm 0.19$
& $0.34 \pm 0.19$ \\

Velocity only
& $0.07 \pm 1.48$
& $-0.38 \pm 0.10$
& $0.55 \pm 0.65$
& $0.08 \pm 0.74$
& $-0.88 \pm 0.30$
& $-0.77 \pm 1.38$
& $0.55 \pm 0.15$
& $-0.37 \pm 0.61$ \\

Point cloud
& $0.54 \pm 1.17$
& $0.53 \pm 0.63$
& $0.24 \pm 0.45$
& $0.44 \pm 0.75$
& $-0.79 \pm 0.08$
& $-1.33 \pm 2.05$
& $-0.90 \pm 0.99$
& $-1.01 \pm 1.04$ \\

Top/Bottom 20\%
& $0.60 \pm 0.61$
& $-0.03 \pm 0.95$
& $0.51 \pm 0.78$
& $0.36 \pm 0.78$
& $-0.01 \pm 0.36$
& $0.66 \pm 0.14$
& $0.53 \pm 0.12$
& $0.39 \pm 0.21$ \\

RGB only
& $0.14 \pm 0.12$
& $-0.74 \pm 4.36$
& $-0.03 \pm 0.12$
& $-0.21 \pm 1.53$
& $-0.06 \pm 0.41$
& $-1.95 \pm 1.34$
& $-0.37 \pm 0.41$
& $-0.80 \pm 0.72$ \\

Masked Depth + ViT-Small
& $-1.77 \pm 0.30$
& $-0.94 \pm 10.73$
& $-0.06 \pm 0.85$
& $-0.92 \pm 3.96$
& $-0.75 \pm 0.14$
& $-0.33 \pm 0.39$
& $-0.47 \pm 0.41$
& $-0.52 \pm 0.31$ \\

\hline
\end{tabular*}
\end{table*}
Table~\ref{tab:force_results} and Table~\ref{tab:r2_results} summarize the results on the rubber-glove phantom and \textit{ex vivo} porcine colon. The proposed pipeline achieves average RMSEs of 0.77~N on the phantom and 1.30~N on the colon, which are the lowest among all methods. It also achieves the highest average $R^2$ values  of 0.71 and 0.57. 
Compared with the proposed method, the RGB-only and masked depth + ViT-Small baselines increase the average RMSE from 0.77~N to 1.60~N and 1.97~N on the phantom, and from 1.30~N to 2.74~N and 2.68~N on the colon. Example force trajectories in Fig.~\ref{fig:force_curve} show that the predicted forces follow the ground-truth trends in both settings.

We visually inspected the estimated object frames, and all orientations were consistent with the tissue shape. Among the ablations, the camera-frame setting causes the largest RMSE increase on the phantom and the lowest average $R^2$ on both datasets. 
The degradation is smallest along the $x$-axis.
The $F_x$ RMSE increases from 0.91~N to 1.31~N on the phantom and from 1.04~N to 1.50~N on the colon. In comparison, the $F_z$ RMSE increases from 0.78~N to 2.17~N and from 2.23~N to 5.02~N, respectively. On the phantom, the position-only and velocity-only settings cause the next largest degradations, increasing the average RMSE to 1.39~N and 1.32~N, with average $R^2$ values of 0.13 and 0.08. On the colon, they increase the average RMSE to 1.74~N and 2.02~N, with average $R^2$ values of 0.34 and $-0.37$.

Tracking strategy also affects force accuracy. Compared with the proposed tracking, CoTracker achieves a lower coverage and NNI on both the phantom and the colon (Table~\ref{tab:tracking_quality}). However, on the colon, the coverage of the two methods is similar. With CoTracker, the average RMSE increases from 0.77~N to 0.96~N on the phantom and from 1.30~N to 1.74~N on the colon. The average $R^2$ decreases from 0.71 to 0.58 and from 0.57 to 0.16, respectively. As shown in Fig.~\ref{fig:cotracker}, the CoTracker points cluster toward the two ends of the tissue, which resembles the point distribution of the top/bottom 20\% setting. The point-cloud setting reaches 1.12~N on the phantom. On the colon, it has the highest average RMSE of all methods (3.07~N) and an average $R^2$ of $-1.01$.

For context, Table~\ref{tab:previous_work} compares our ex vivo colon results with published robot-state-based and vision-only estimates. Experimental conditions differ across studies, and the vision-only results in~\cite{Yang2024} include ground-truth-based rescaling.

\begin{table}[htbp]
\centering
\caption{Comparison with previously reported force estimation
results. Values are averaged across the three force components.
Experimental conditions differ across studies.}
\label{tab:previous_work}

\footnotesize
\renewcommand{\arraystretch}{1.1}
\setlength{\tabcolsep}{3pt}

\begin{tabular*}{\columnwidth}
{@{\extracolsep{\fill}}llcc@{}}
\toprule
\textbf{Method} &
\textbf{Setting} &
\textbf{RMSE (N)} &
\textbf{NRMSE (\%)} \\
\midrule
Yang et al.~\cite{Yang2025}
& dVRK-Si, Rigid shaft & 2.16 & 5.21 \\
Yang et al.~\cite{Yang2024}
& dVRK, Chicken tissue & -- & 10.5 \\
Proposed
& dVRK-Si, Porcine colon & 1.30 & 9.12 \\
\bottomrule
\end{tabular*}
\end{table}

In the \emph{in vivo} sequence (Fig.~\ref{fig:invivo_validation}), the predicted 3D forces change consistently with the observed direction and extent of tissue deformation: the force responses increase as the operator pulls the tissue. On an NVIDIA RTX A5000, depth estimation runs at 1 FPS, SAM2 at 24 FPS, tracking at 26 FPS, and force prediction at 37,000 FPS, limiting the end-to-end rate to about 1 FPS.

\section{Discussion}
Our results demonstrate the feasibility of estimating 3D interaction forces from tissue deformation reconstructed using stereo endoscopic video. Among the evaluated components, object-frame alignment has the largest effect on accuracy. Because deformation is represented in a more consistent way, the model does not need to learn variations caused by tissue rotation and changes in the ECM pose. Camera-frame and image-based inputs require the model to infer tissue orientation implicitly. The $x$-axis shows a relatively small degradation between the object-frame and camera-frame settings. One possible reason is that the $x$-axis is approximately aligned with the vertical direction in our experimental setup and is therefore less affected by horizontal rotation of the tissue. 
Point velocity also provides important information for force estimation. Position describes the current tissue shape, and velocity captures the direction and rate of deformation. This provides additional information about the temporal evolution of the interaction. Removing velocity therefore causes a consistent performance drop in both the phantom and \emph{ex vivo} experiments.

The tracking experiments further highlight the importance of reliable temporal correspondence. 
As illustrated in Fig.~\ref{fig:cotracker}, CoTracker loses many tracks in the central tissue region, while the remaining points are mainly distributed near the two ends of the longitudinal axis. Its spatial distribution is therefore similar to the top/bottom 20\% ablation, which also achieves comparable performance on the colon. This observation suggests that, for pulling tasks with partial occlusion, temporal motion and relative deformation of the tracked regions may be more important than maintaining dense coverage over the entire tissue surface. In contrast, CoTracker is less stable on the glove phantom, where several tracking points cluster into small regions. The full point-cloud input performs poorly on the ex vivo data. The reconstructed ex vivo surface is geometrically more complex, while the point-cloud-only representation does not maintain fixed point correspondences across frames, making point-wise velocity unavailable.

Compared with the robot-state-based dVRK-Si method in~\cite{Yang2025}, our colon results show a lower absolute RMSE (1.30 versus 2.16~N), but a higher NRMSE
(9.12\% versus 5.21\%), illustrating the influence of force range on these metrics. Our NRMSE is also close to the 10.5\% obtained by the vision-only method
in~\cite{Yang2024}, although that method rescaled its predictions against test ground truth. These comparisons provide context rather than establish superiority across different experiments. Our results demonstrate force estimation from stereo video without instrument kinematics on ex vivo colon under held-out tissue orientations and camera poses.

The qualitative \emph{in vivo} evaluation in Fig.~\ref{fig:invivo_validation} further suggests that the learned deformation representation can transfer beyond the controlled settings. Although no synchronized force measurements are available, the qualitative agreement between tissue deformation and predicted force suggests that the learned representation is not completely tied to the appearance or configuration of the training scenes.

There are a few limitations. First, the object frame estimation depends on the tissue shape and the ECM pose. The ECM pose is used only to align the object-frame $X$-axis with the vertical force-sensor $X$-axis, so that predictions and ground truth share a frame. Future work will evaluate fully vision-derived alternatives such as image-based reference. In all phantom and ex vivo sequences, the full tissue is visible in the reference frame. We did not evaluate frame estimation when only part of the tissue is visible. In realistic surgical scenes, such partial views can make the global orientation ambiguous. Additionally, as our focus is on colon anastomosis, our object-frame definition assumes a long, tubular object. Anatomy-specific geometric priors could provide a more robust frame definition. Second, the current pipeline relies heavily on stereo depth estimation. Errors from camera calibration, specular reflections, weak texture, tissue folding, and large occlusions can affect the reconstructed point cloud and subsequently the estimated deformation. Depth estimation is also the slowest stage, limiting the end-to-end rate. Accurate stereo calibration and efficient real-time depth estimation are therefore important for deployment. In addition, the geometry-constrained tracking relies on SAM2 to determine point visibility. 
The estimated object-frame origin has a small positional offset from the force-sensor origin. Since this work estimates force rather than torque, this offset does not affect the force components when the two frames are aligned in orientation. Future work could use stronger geometric priors to estimate the tissue--sensor frame relationship more accurately. 
The current force model is also trained on a limited range of tissue mechanical properties and experimental configurations. Variations in tissue stiffness, thickness, fixation, and pre-tension can change the deformation--force relationship. Future experiments should therefore include a wider range of tissue types and mechanical conditions to improve generalization.

\section*{Conclusion}

In this work, we present a vision-based framework for estimating 3D interaction forces from stereo endoscopic video. The framework combines object-centered tissue reconstruction, geometry-constrained point tracking, and a neural network model to predict force from tissue deformation. Across held-out tissue poses and camera viewpoints, it achieved average RMSEs of 0.77~N on rubber-glove phantoms and 1.30~N on \emph{ex vivo} porcine colons, outperforming the evaluated baselines. Ablation studies supported the contributions of object-frame alignment, temporal tracking, and point velocity. In an \emph{in vivo} colorectal surgical sequence, the predicted forces were qualitatively consistent with the observed tissue manipulation. Future work will focus on improving reconstruction under occlusion, evaluating a wider range of tissue properties, and quantitatively validating force estimates in vivo.

\section*{Acknowledgment}
We used ChatGPT and Claude for language editing and assisting part of the data preprocessing and visualization code. All results were verified by the authors. This study was approved by the relevant Institutional Review Board (IRB). All identifying information was removed from the clinical surgical video before analysis.


\bibliographystyle{IEEEtran}
\bibliography{refs}

\end{document}